\documentclass[conference]{IEEEtran}
\IEEEoverridecommandlockouts
\usepackage{enumitem}
\usepackage{cite}
\usepackage{amsmath,amssymb,amsfonts}
\usepackage{algorithmic}
\usepackage{ dsfont }
\usepackage{pifont}%
\usepackage{graphicx}
\usepackage{textcomp}
\usepackage{xcolor}
\usepackage{xspace}
\usepackage{multirow}
\usepackage{booktabs}
\usepackage{adjustbox}
\usepackage{rotating}
\newcommand{\Design}{\textit{$\mathsf{TACO}$\xspace}}
\def\BibTeX{{\rm B\kern-.05em{\sc i\kern-.025em b}\kern-.08em
    T\kern-.1667em\lower.7ex\hbox{E}\kern-.125emX}}
\begin{document}

\title{Transformer-Based Contrastive Meta-Learning For Low-Resource Generalizable Activity Recognition\vspace{-3mm}}

\author{
    \IEEEauthorblockN{Junyao Wang$^\dag$, Mohammad Abdullah Al Faruque$^\dag$$^\S$}
    \IEEEauthorblockN{\textit{$^\dag$ Department of Computer Science, University of California, Irvine, CA, United States}\\
    $^\S$ \textit{Department of Electrical Engineering and Computer Science, University of California, Irvine, CA, United States}
    \\\textit{\{junyaow4, alfaruqu\}}@uci.edu\vspace{-3mm}}
}

\maketitle

\begin{abstract}
Deep learning has been widely adopted for human activity recognition (HAR) while \textit{generalizing} a trained model across diverse users and scenarios remains challenging due to \textit{distribution shifts} (DS). 
The inherent \textit{low-resource} challenge in HAR, i.e., collecting and labeling adequate human-involved data can be prohibitively costly, further raising the difficulty of tackling DS. 
We propose $\Design$, a novel \underline{t}r\underline{a}nsformer-based \underline{co}ntrastive meta-learning approach for generalizable  HAR.
$\Design$ addresses DS by synthesizing 
virtual target domains in training with explicit consideration of model generalizability. 
Additionally, we extract expressive feature with the attention mechanism of Transformer and incorporate the supervised contrastive loss function within our meta-optimization to enhance representation learning.
Our evaluation demonstrates that $\Design$ achieves notably better performance across various low-resource DS scenarios. 
\end{abstract}


\section{Introduction}
Human activity recognition (HAR), 
using sensor data to identify user activities, is essential for advancing healthcare and improving human well-being~\cite{kddgeneralizable}. 
Deep neural networks have been widely adopted for time-series data processing~\cite{hammerla2016deep,wang2023robust,wang2023disthd} and notably improved HAR over traditional methods~\cite{bulling2014tutorial, hammerla2015pd}. 
However, model \textit{generalizability} remains a critical challenge due to \textit{distribution shifts} (DS) caused by user and configuration discrepancies, e.g., different living habits and sensor positions. 
Existing HAR models rely on the strong assumption that training and inference samples come from the same data distribution while this can be easily violated in practice and cause model failures~\cite{cheng2020skeleton, wang2024rs2g}.
For instance, a model trained on data from existing patients can seriously fail when tested on new patients in different body status~\cite{wang2023domino,wang2024smore}.
Additionally, existing deep learning-based HAR require extensive labeled data to achieve reasonable performance while human-involved data are privacy-sensitive,  costly to annotate, and susceptible to variations of sensors and environments~\cite{saeed2019multi,wang2023hyperdetect}. 
The \textit{low-resource} data hardly represent the diverse human activities and heightens the difficulty of developing generalizable models.

To address these challenges, we propose $\Design$, a novel transformer-based contrastive meta-learning approach for the domain generalization (DG) of HAR. 
$\Design$ learns domain-invariant and class-discriminative representations that can be effectively generalized to unseen instances in target domains with limited training samples. 
We mitigate the low-resource challenge in HAR by expanding the diversity of the data space with specially designed sensor data augmentations and extracting expressive features leveraging the attention mechanism of Transformer. 
 We then simulate DS during the training of meta-learning by synthesizing virtual target domains within each mini-batch. 
Our meta-optimization objective explicitly considers model \textit{generalizability}, requiring optimization steps that improve the performance in meta-train domains simultaneously improve the performance in virtual target domains.
We incorporate a supervised contrastive loss function within our meta-optimization to further enhance representation learning. Our  contributions are listed as follows:
\begin{itemize}[leftmargin = *]
    \item We propose $\Design$, a novel robust and generalizable HAR algorithm that consistently provides high-quality performance across various low-resource distribution shift scenarios. $\Design$ exhibits on average $4.08\%$ higher accuracy than state-of-the-art domain generalization approaches.  
    \item To the best of our knowledge, $\Design$ is the first HAR algorithm with supervised contrastive learning incorporated in meta-learning.   
    Our meta-optimization explicitly considers domain generalizability, wherein the supervised contrastive loss function enhances semantic discrimination of features.
\item $\Design$ utilizes well-designed data augmentation techniques to expand the diversity of data space and employs the attention mechanism of Transformer to enhance the representation learning of multivariate time series data, effectively alleviating the demand for massive labeled data. 
\end{itemize}
\section{Methodology}

\begin{figure*}[t] 
\centering
    \includegraphics[width=1.\textwidth]{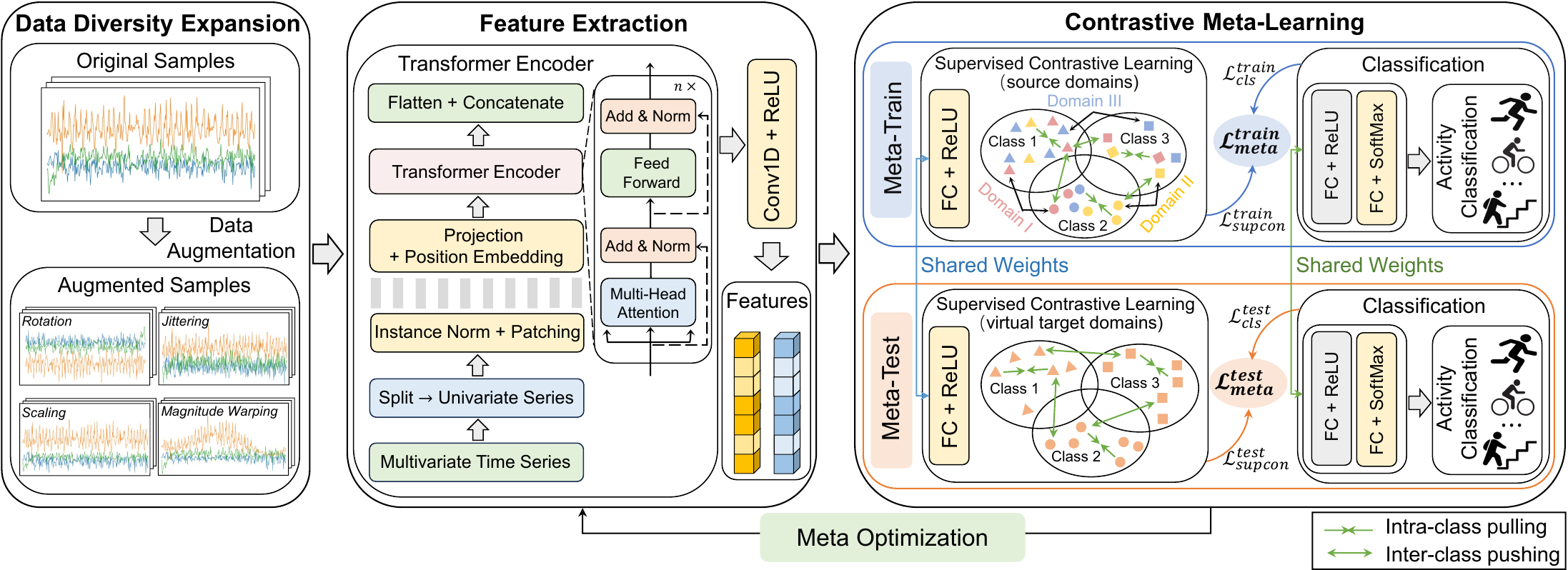}
    \vspace{-4mm}
   \caption{The architecture overview of $\Design$. Different colors in the feature space indicate different domains. }
    \label{fig:train}
  \vspace{-2mm}
\end{figure*}

\subsection{Problem Formulation}
A domain is defined as a joint probability distribution $ P_{(X, Y)}$ on $X \times Y$, where $X$ and $Y$ denote the instance space and label space, respectively.
Our goal is to learn a classification model $f: X \rightarrow  Y$ with samples in source domains $S$ to capture domain-invariant representations so that $f$ can be generalized to new instances from \textit{unseen} target domains $T$.
Note that only information from source domains can be used for training, i.e., data from target domains cannot be accessed until deployment.  
DS is a prevalent issue in HAR due to the diversity of users and configurations.
It arises when a target domain has notably different joint probability distribution from the source domains, i.e., $P^ T_{(X,Y)}  \neq   P^{S}_{(X,Y)}$, 
 so that a model trained on source domains can fail to adapt to instances from target domains. 
Additionally, given the low-resource challenge in HAR, source domains often consist of limited training samples, further increasing the difficulty of developing a robust and generalizable model to tackle DS.

\subsection{Data Diversity Expansion} \label{subsec:augment}
We expand data diversity with the following data augmentations motivated by ~\cite{saeed2019multi,khaertdinov2021contrastive,kddgeneralizable}, explicitly considering temporal properties and motion patterns in sensor-based time series data. 

\begin{itemize}[leftmargin = *]
\itemsep0em 
    \item \textbf{Rotation:} Rotate data to simulate sensor orientations on human bodies to achieve sensor-placement invariance 
    \item \textbf{Permutation:} Slice a window of data into segments and randomly permute them to form a new window. This helps to explore permutation-invariant properties in learning. 
    \item \textbf{Scaling:} Scale the magnitude of window-length data to enhance robustness against amplitude and offset variances. 
    \item \textbf{Time Warping:} Perturb the temporal location by smoothly distorting the time intervals between samples. This broadens local diversity in the temporal dimension. 
    \item \textbf{Magnitude Warping:} Convolve data window with a smooth curve to broaden magnitude diversity. 
    \item \textbf{Jittering:} Add random noise to samples, enabling the model to capture features that are invariant to minor corruption. 
\end{itemize}

\subsection{Transformer-Based Feature Extraction}\label{subsec:transformer}
Transformer captures connections between elements in a sequence and is considered an ideal approach for extracting local semantic information and analyzing connections among time steps. 
We leverage the patch time series Transformer (PatchTST) encoder~\cite{nie2022time} to capture comprehensive semantic information. 
We segment time series data into subseries-level patches and feed them to Transformer as input tokens. 
As shown in Fig. \ref{fig:train}, we consider a collection of multivariate time series samples with look-back window $L: ({\bf{x}}_1, \ldots, {\bf{x}}_L)$, where each ${\bf{x}}_t$ at time step $t$ $(t= 1, \ldots, L)$ is a vector of dimension $M$ representing values from $M$ sensors. 
We split the multivariate input $({\bf{x}}_1, \ldots, {\bf{x}}_L)$ into $M$ univariate series, and denote the univariate series from the $i$-th sensor as ${{\bf{x}}}^{(i)}=(x_1^{(i)}, \ldots, x_{L}^{(i)})$, where $i=1, \ldots, M$.
The univariate time series ${\bf{x}}^{(i)}$ is then divided into patches, generating a sequence of patches ${\bf{x}}^{(i)}_{p} \in \mathds{R}^{ L' \times  N}$, where $L' (L'\leq L)$ is patch length and  $N$ is the number of patches. 
We utilize a vanilla Transformer encoder to map the patches to the latent space of dimension $D$ via a trainable linear projection ${\bf{W}}_p\in \mathds{R}^{ D\times  L'}$, and apply a learnable additive position encoding ${\bf{W}}_{pos}\in \mathds{R}^{D\times  N}$ to monitor the temporal order of patches. 
The input that will be fed into Transformer encoder is calculated as ${\bf{x}}_d^{(i)}={\bf{W}}_p{\bf{x}}^{(i)}_{p}+{\bf{W}}_{pos}$. 
Each head $h=1, \ldots, H$ in multi-head attention will transform ${\bf{x}}_d^{(i)}$ into query matrices ${Q}_h^{(i)}=({\bf{x}}_d^{(i)})^T{\bf{W}}_{h}^{ Q}$, key matrices ${K}_h^{(i)}=({\bf{x}}_d^{(i)})^T{\bf{W}}_{h}^  K$ and value matrices ${V}_h^{(i)}=({\bf{x}}_d^{(i)})^T{\bf{W}}_{h}^ V$, where ${\bf{W}}_{h}^Q, {\bf{W}}_{h}^K\in \mathds{R}^{D \times d_k}$ and ${\bf{W}}_{h}^V\in \mathds{R}^{D \times D}$, and $d_k$ is the dimensionality of the key vectors. 
Then a scaled production is used to get attention output:
\begin{equation*}
    \textit{Attention}({Q}_h^{(i)}, {K}_h^{(i)}, { V}_h^{(i)})=\textit{Softmax}\Big(\frac{{\mathcal Q}_h^{(i)} {K}_h^{{(i)}^T}}{\sqrt{d_k}}\Big){V}_h^{(i)},
\end{equation*}
The multi-head attention block also includes BatchNorm layers and a feed-forward network with residual connections as shown in Fig. \ref{fig:train}. Transformer encoder generates representations ${\bf{z}}^{(i)}\in \mathds{R}^{D\times  N}$. 
We flatten and concatenate the transformed sequence ${\bf{z}}^{(1)}, {\bf{z}}^{(2)}, \ldots, {\bf{z}}^{(M)}$ into ${\bf{z}}\in \mathds R^{M\times {DN}}$, followed by a 1D-convolutional layer for feature extraction.

\subsection{Contrastive Meta-Learning}  \label{subsec:contrast_meta}
As demonstrated in Fig. \ref{fig:train}, we propose a novel contrastive meta-learning approach to enhance representation learning and improve model generalizability. 
Motivated by~\cite{li2018learning}, we 
simulate DS in the training of meta-learning by synthesizing virtual testing domains in each iteration. 
We split $U$ source domains into $(U - V)$ meta-train and $V$ meta-test domains (virtual target domains) to mimic real DS. 
Our meta-optimization objective requires optimization steps improving performance in meta-train domains simultaneously improve performance in virtual target domains.
Our meta-loss function incorporates supervised contrastive loss~\cite{khosla2020supervised} with task-oriented classification loss to further ensure accurate HAR.

\subsubsection{Supervised Contrastive Loss}
We employ the supervised contrastive loss function~\cite{khosla2020supervised} nested within our meta-optimization to enhance the semantic discrimination of representations. 
We enlarge the inter-class distance, i.e., the distance between samples from different classes, and minimize the intra-class distance, i.e., the distance between samples from the same class for all the original and augmented samples. 
We randomly regard a sample as the \textit{anchor}, and consider all the other samples in the same class \textit{positive} of the anchor while all the others \textit{negative} of the anchor. Mathematically,
\begin{equation*}
    \mathcal L_{supcon}=\sum_{i\in \mathcal I}\frac{-1}{|\mathcal P(i)|}\sum_{p\in \mathcal P(i)}\log{\frac{\exp(z_i\cdot z_p/\tau)}{\sum_{a\in \mathcal A(i)}\exp({z_i\cdot z_a/\tau})}}
\end{equation*}
where $\mathcal I$ is the index set of the original and the augmented representations and the index $i$ denotes the anchor.
$\mathcal A(i)\equiv \mathcal I\setminus \{i\}$ is all the samples excluding the anchor, $\mathcal P(i)\equiv \{p\in \mathcal A(i): \tilde{y_p} = \tilde{y_i}\}$ is the set of indices of the representations with the same label as the anchor, 
$z_p$ is the positive representation, and $\tau \in \mathds R^+$ is the scalar temperature.
Note that we only consider samples from the $( U-  V)$ meta-train domains for meta-train and samples from the $V$ virtual target domains for meta-test. 
Minimizing the distance between the original and augmented samples enables the model to capture intrinsic representations and become more robust against real-world DS. 
Additionally, with the presence of labels, supervised contrastive learning enhances the model's ability to learn class-discriminative representations and deliver accurate HAR. 
 
\subsubsection{Task-Oriented Classification Loss}
We utilize fully connected layers for human activity classification. 
We learn the model $f:  X_{all}\rightarrow  Y$, where $X_{all}$ includes the original and the augmented data 
and $Y$ denotes the label space. 
We calculate the classification loss with cross-entropy, i.e.,
\begin{equation*}
    \mathcal L_{cls}=-\sum_{i=1}^{ C} y_i \log \hat{y_i}
\end{equation*}
where $ C$ denotes the number of classes, $\hat{y_i}$ denotes the output of the linear classifier for the $i$-th class and $y_i$ denotes the true binary indicator for the $i$-th class. 

\subsubsection{Meta Optimization}
Our meta-optimization loss function is defined as the combination of the supervised contrastive loss and the task-oriented classification loss, i.e., 
\begin{equation}\label{eq:meta}
    \mathcal L_{meta}(\cdot) =  \eta \cdot \mathcal L_{cls}(\cdot) + (1-\eta) \cdot \mathcal L_{supcon}(\cdot)
\end{equation}
where $\eta$ is a hyper-parameter controlling the relative impact of the task-oriented classification loss and the supervised contrastive loss. 
Denoting the meta-train domains as $S_1, S_2, \ldots, S_{U-V}$ and the model parameters as $\Theta$,
in the meta-train stage, the model is updated on the meta-loss calculated with all the $(U - V)$ meta-train domains, i.e., 

\begin{equation}
    \mathcal L^{train}_{meta}(\Theta)=\frac{1}{U - V}\sum_{i=1}^{U - V}\frac{1}{|S_{i}|}\sum_{j=1}^{|S_{i}|}\mathcal L^{(j)}_{meta}(\Theta)
\end{equation}
where $|S_{i}|$ denotes the number of samples in meta-train domain $i$ $(1\leq i\leq S - V)$. 
The meta-train stage updates the model with a learning rate $\alpha$, which can be formulated as 
\begin{equation*}
    \Theta' = \Theta - \alpha \cdot \frac{\partial \mathcal L^{train}_{meta}(\Theta)}{\partial \Theta}
\end{equation*}
The updated model is then evaluated on $V$ virtual target domains in the meta-test stage, simulating testing on new domains with different data distributions. 
Denoting the virtual target domains as $T_1, T_2, \ldots, T_\mathcal V$, 
the loss of adapted parameters on the virtual target domains can be calculated as 

\begin{equation}
    \mathcal L^{test}_{meta}(\Theta') = \frac{1}{V}\sum_{i=1}^{V}\frac{1}{|T_{i}|}\sum_{j=1}^{| T_{i}|}\mathcal L_{meta}^{(j)}({\Theta'})
\end{equation}
where $|T_{i}|$ $(1\leq i\leq V)$ is the number of samples in the virtual target domain $T_i$. Note that the loss on the virtual target domain is calculated with the updated parameters $\Theta'$ from meta-train. 
To enable optimization in 
both the meta-train and virtual target domains, the final loss function is defined as:  
\begin{equation}\label{eq:final}
    \mathcal L(\Theta) =  \mathcal L^{train}_{meta}(\Theta)+\beta\cdot \mathcal L^{test}_{meta}\Big(\Theta-\alpha\cdot \frac{\partial \mathcal L^{train}_{meta}}{\partial \Theta}\Big)
\end{equation}

\section{Experimental Result}
\subsection{Setup}

We evaluate $\Design$ on popular public HAR datasets including DSADS~\cite{barshan2014recognizing}, PAMAP2~\cite{reiss2012introducing}, and USC-HAD~\cite{zhang2012usc}. 
To construct DS scenarios, we randomly divide subjects into groups for leave-one-domain-out (LODO) evaluation. 
For DSADS and PAMAP2, 
we divide 8 subjects into 4 groups, each consisting of data from 2 subjects. 
For USC-HAD, we 
divide 14 subjects into 5 groups, where each of the first three groups contains 3 subjects and the last group contains 2 subjects. 
For LODO evaluation, we consider one group of subjects as the target domain and the remaining groups as source domains. 
In meta-learning, we split source domains into meta-train domains and virtual target domains in a similar LODO manner. 
To simulate low-resource scenarios in HAR, 
we construct training data by randomly sampling $20\%$ to $100\%$ data from source domains with a step of $20\%$, and evaluate the trained model on the target domain to show the robustness of $\Design$ across various low-resource scenarios.
\begin{table*}[ht]
\vspace{-1mm}
\begin{adjustbox}{width=\textwidth,center}
    \centering
    \begin{tabular}{lc| c c c c c c c c c c} \toprule 
          &\begin{tabular}[c]{@{}c@{}}Target\\ Domain\end{tabular}  & \begin{tabular}[c]{@{}c@{}}ERM\\ \footnotesize{\cite{vapnik1991principles}}\end{tabular} & 
          \begin{tabular}[c]{@{}c@{}}MLDG\\ {\footnotesize\cite{li2018learning}}\end{tabular} & \begin{tabular}[c]{@{}c@{}}RSC\\ \footnotesize{\cite{huang2020self}}\end{tabular} & \begin{tabular}[c]{@{}c@{}}AND-Mask\\ \footnotesize{\cite{parascandolo2020learning}}\end{tabular} & \begin{tabular}[c]{@{}c@{}}SimCLR\\ \footnotesize{\cite{chen2020simple}}\end{tabular} & \begin{tabular}[c]{@{}c@{}}Fish\\ 
          \footnotesize{\cite{shi2021gradient}}\end{tabular} & 
           \begin{tabular}[c]{@{}c@{}}DOMINO\\ \footnotesize{\cite{wang2023domino}}\end{tabular} &
          \begin{tabular}[c]{@{}c@{}}DDLearn\\ \footnotesize{\cite{kddgeneralizable}}\end{tabular} & \begin{tabular}[c]{@{}c@{}}$\Design$\\ \footnotesize{(ours)}\end{tabular}\\ \midrule
        \multirow{5}{0.05em}{\begin{turn}{90}DSADS\end{turn}}  
        &T0& $59.53(\pm3.17)$   & $64.14(\pm3.79)$ & $60.12(\pm1.96 )$ & $60.35(\pm1.43)$ &  $72.48(\pm3.18)$ & $57.53(\pm0.92)$ &$60.15(\pm1.03)$&  \underline{$87.88(\pm 1.92)$} & ${\bf{93.97}}(\pm1.06)$\\  
          &T1& $64.21(\pm 6.02)$   &  $69.77(\pm2.93)$  & $72.93(\pm7.76)$ & $65.30(\pm2.02)$ & $76.61(\pm2.56)$ & $65.45(\pm1.18)$& $64.70(\pm1.38)$  & \underline{$88.80(\pm 1.11)$} & ${\bf{92.24}}(\pm1.63)$  \\ 
          &T2& $66.91(\pm 5.28)$  &  $72.73(\pm0.98)$  & $78.87(\pm3.17)$ & $69.21(\pm2.18)$ & $78.25(\pm0.92)$ & $74.62(\pm 2.13)$ & $72.26(\pm0.72)$ &  \underline{$89.21(\pm1.23)$} & ${\bf{91.77}}(\pm0.78)$  \\  
          &T3& $65.16 (\pm 6.21)$    &  $73.14(\pm1.26)$ &  $72.15(\pm5.72)$  & $70.29(\pm2.50)$ & $76.49(\pm0.91)$ & $67.28(\pm1.42)$& $71.35(\pm2.04)$ & \underline{$85.63(\pm1.13)$} & ${\bf{90.46}}(\pm1.52)$\\ 
          &Avg&  $63.95(\pm5.17)$ &     $69.95(\pm2.24)$ &  $71.02(\pm4.65)$  & $66.29(\pm2.03)$ & $75.96(\pm1.90)$ & $66.22(\pm1.41)$& $67.12(\pm1.30)$  & \underline{$87.88(\pm1.35)$} & ${\bf{92.11}}(\pm1.25)$ \\  \midrule
         \multirow{5}{0.05em}{\begin{turn}{90}PAMAP2\end{turn}} 
         &T0& $50.03(\pm2.91)$   &    $60.45(\pm6.14)$ &  $64.18(\pm3.73)$  & $60.76(\pm4.38)$ & $63.28(\pm3.33)$ & $64.90(\pm3.44)$& $62.64(\pm2.70)$  & \underline{$75.55(\pm0.79)$} & ${\bf 79.39}(\pm1.92)$ \\  
          &T1& $80.37(\pm7.83)$    &    $79.91(\pm3.87)$ &  $87.45(\pm6.17)$  & $85.29(\pm4.70)$ & $81.25(\pm1.59)$ & $82.06(\pm2.45)$& $79.95(\pm2.19)$  & \underline{${{90.07}}(\pm2.40)$} & ${\bf{92.45}}(\pm1.37)$\\ 
          &T2&  $76.12(\pm2.95)$    &    $78.68(\pm3.49)$ &  $76.60(\pm2.26)$  & {$81.90(\pm2.54)$} & $78.65(\pm1.87)$ & $77.38(\pm2.58)$& $81.36(\pm1.92)$  & ${\bf{85.51}}(\pm0.76)$ & $\underline{84.46(\pm1.15)}$\\ 
          &T3& $65.81(\pm5.92)$   &    $75.73(\pm1.62)$ &  $74.65(\pm4.93)$  & $72.85(\pm3.22)$ & $71.09(\pm1.99)$ & \underline{$82.36(\pm3.89)$}& $79.68(\pm2.05)$  & $80.67(\pm1.78)$ & ${\bf{85.98}}(\pm 1.71)$\\ 
          & Avg& $68.08(\pm4.90)$     &    $73.69(\pm3.78)$ &  $75.72(\pm4.27)$  & $75.20(\pm3.71)$ & $73.57(\pm2.20)$ & $76.68(\pm3.09)$& $75.91(\pm2.22)$  & \underline{$82.95(\pm1.43)$} & ${\bf{85.57}}(\pm1.54)$\\ \midrule
         \multirow{5}{0.05em}{\begin{turn}{90}USC-HAD\end{turn}} 
         &T0& $72.57(\pm2.77)$    &    $70.10(\pm2.57)$ &  \underline{$80.30(\pm3.14)$}  & $72.66(\pm3.94)$ & $69.36(\pm2.34)$ & $79.44(\pm2.83)$& $76.27(\pm1.48)$  & $79.06(\pm2.11)$ & ${\bf{85.94}}(\pm1.29)$\\  
          &T1& $70.28(\pm1.74)$    &    $68.22(\pm1.92)$ &  $77.23(\pm1.97)$  & $71.26(\pm1.25)$ & $66.62(\pm1.44)$ & $75.39(\pm1.56)$&\underline{$80.64(\pm2.13)$}   & $80.15(\pm1.11)$ & ${\bf{86.22}}(\pm0.74)$\\ 
          &T2& $69.86(\pm1.90)$ &    $71.03(\pm4.32)$ &  $75.15(\pm4.26)$  & $70.44(\pm1.76)$ & $76.04(\pm1.61)$ & $73.85(\pm2.17)$ &$77.91(\pm1.66)$ &  \underline{$80.81(\pm0.74)$} & ${\bf{82.15}}(\pm0.93)$ \\ 
          &T3& $57.72(\pm6.26)$  &    $59.75(\pm5.96)$ &  \underline{$71.24(\pm5.46)$}  & $65.05(\pm6.27)$ & $61.24(\pm1.06)$ & $62.03(\pm2.99)$& $66.30(\pm3.53)$  & $70.93(\pm1.87)$ & ${\bf{78.64}}(\pm1.59)$\\ 
          &T4& $64.26(\pm2.31)$  &    $65.94(\pm0.79)$ &  $68.90(\pm2.93)$  & $63.57(\pm2.24)$ & $62.85(\pm2.17)$ & $61.98(\pm6.24)$& $68.16(\pm1.84)$  & \underline{$75.87(\pm2.99)$} & ${\bf{80.85}}(\pm1.77)$\\ 
          & Avg& $66.94(\pm3.00)$ &    $67.00(\pm3.11)$ &  $74.56(\pm4.44)$  & $68.60(\pm3.10)$ & $67.22(\pm1.72)$ & $70.54(\pm3.16)$& $73.86(\pm2.13)$  & \underline{$77.36(\pm1.76)$} & ${\bf{82.76}}(\pm1.26)$\\ 
          \bottomrule
    \end{tabular}
    \end{adjustbox}
    \vspace{1mm}
    \caption{Classification accuracy ($\%$)($\pm$ standard deviation) in low-resource setting with only $20\%$ of the training data. The best and the second-best results are marked in \textbf{bold} and \underline{underlined}, respectively. }
    \label{tab:overall_acc}
  \vspace{-3mm}
\end{table*}


\begin{table*}[ht]
\begin{adjustbox}{width=\textwidth,center}
    \centering
    \begin{tabular}{lc| c c c c c c c c c c} \toprule 
          &\begin{tabular}[c]{@{}c@{}}Training\\ Proportion\end{tabular}  & ERM~ \footnotesize{\cite{vapnik1991principles}} & 
          MLDG~{\footnotesize\cite{li2018learning}} & RSC~{\cite{huang2020self}} & AND-Mask~ \footnotesize{\cite{parascandolo2020learning}} & SimCLR~\footnotesize{\cite{chen2020simple}} & Fish~\footnotesize{\cite{shi2021gradient}} & 
           DOMINO~ \footnotesize{\cite{wang2023domino}} &
          DDLearn ~\footnotesize{\cite{kddgeneralizable}} & $\Design$  \footnotesize{(ours)}\\ \midrule
        \multirow{5}{0.05em}{\begin{turn}{90}DSADS\end{turn}}  
        &20\%& $63.95$ & $69.88$ & $71.02$& $68.29$& $75.96$ & $66.22$&$67.12$  & \underline{$87.88$} & ${\bf92.11}$\\
          &40\%& $65.39$  & $72.24$ & $73.18$ & $70.07$&$75.76$ & $69.73$& $69.77$&\underline{$89.71$} & ${\bf93.57}$\\
          &60\%& $68.17$ & $75.08$ & $77.64$ & $72.21$&  $75.61$ &$71.57$ & $71.24$&\underline{$90.43$}& ${\bf93.89}$\\  
          &80\%&  $71.08$ & $75.65$ & $78.05$& $74.92$&  $76.69$ & $73.96$& $73.92$&\underline{$90.97$}& ${\bf94.02}$\\
          &100\%&  $73.51$   & $76.19$ & $78.22$& $76.13$& $77.24$ &$75.64$ & $75.73$&\underline{$91.95$}& ${\bf94.51}$\\
         \midrule
         \multirow{5}{0.05em}{\begin{turn}{90}PAMAP2\end{turn}} 
         &$20\%$& $68.08$  & $73.69$ &$75.72$ & $75.20$ & $73.57$ & $75.68$  & $75.91$ & \underline{$82.95$} & ${\bf85.57}$\\  
          &$40\%$& $72.87$ & $76.97$ & $79.04$&$76.92$ & $74.25$&$77.38$& $77.38$&\underline{$84.34$}& ${\bf86.46}$\\ 
          &$60\%$&  $74.21$ & $78.14$ & $82.58$& $77.48$& $74.71$& $77.92$& $78.72$&\underline{$86.72$}& ${\bf86.91}$\\ 
          &$80\%$&  $75.95$ & $79.50$  & \underline{$86.73$}&$78.90$ & $76.27$& $78.71$& $79.18$&$\bf{88.92}$& $\underline{87.55}$\\ 
          &$100\%$& $76.33$ & $79.72$ & \underline{$87.86$}&$79.61$ &  $76.88$& $80.29$& $80.09$&\underline{$89.31$}& ${\bf89.48}$\\  \midrule
         \multirow{5}{0.05em}{\begin{turn}{90}USC-HAD\end{turn}} 
         &$20\%$& $66.94$ & $67.00$ & $74.56$ & $68.60$ & $67.22$ & $70.54$ &$73.86$ &  \underline{$77.36$} & ${\bf 82.76}$\\  
          &$40\%$&  $69.17$ & $71.53$ & $76.29$& $72.92$&$69.16$&$74.93$&$74.62$ &\underline{$80.72$}& ${\bf83.49}$\\ 
          &$60\%$&  $72.31$ & $73.07$& \underline{$81.83$}&$74.39$ &$71.38$ &$76.46$& $74.94$&$80.88$& ${\bf83.97}$\\ 
          &$80\%$&  $74.83$ & $75.04$ & $82.10$ & $77.05$&$71.99$&$76.87$&$75.33$ &\underline{$82.49$}& ${\bf84.28}$\\ 
          &$100\%$& $76.10$ & $75.91$ &\underline{$82.94$}& $77.68$&$72.14$&$77.32$& $76.02$&$82.51$& ${\bf86.47}$\\
          \bottomrule
    \end{tabular}
    \end{adjustbox}
    \vspace{2mm}
    \caption{Classification accuracy with partial training data. The best (second-best) are marked in \textbf{bold} (\underline{underlined}) }
    \label{tab:partial_acc}
  \vspace{-5mm}
\end{table*}

\subsection{Network Architecture and Training}  
We conduct experiments on a Linux server with an Intel Xeon Silver 4310 CPU and an NVIDIA GeForce RTX 4090 GPU, 
and reproduce the result of each model following DomainBed~\cite{gulrajani2020search}. 
 $\Design$ is implemented with PyTorch. For our Transformer encoder, denoting the look-back window as $L$ and the patch length as $L'$, we set $L=125$ and $L'=16$ for DSADS, $L=512$ and $L'=64$ for PAMAP2, and $L=500$ and $L'=64$ for USC-HAD.  We set stride as $4$ for DSADS, and as $8$ for PAMAP2 and USC-HAD.
Our Transformer encoder contains 3 layers with the head number $H = 4$, the dimension of latent space $D = 32$, and 
a dropout rate of $0.2$. 
We apply a 1D convolutional layer (kernel size = 8) to process the outputs of Transformer encoder. 
The project head of our supervised contrastive learning is a fully connected layer with 256 neurons, sharing the same set of weights across the meta-train and meta-test. 
Similarly, we use two fully connected layers for task-oriented activity classification, each containing 256 neurons and sharing the same set of weights across the meta-train and meta-test. 
The hyperparameter $\eta$ in equation (\ref{eq:final}) 
is set to 0.2. 
The learning rate $\alpha$ and the parameter $\beta$ in meta-optimization in equation (\ref{eq:meta}), are set to 0.0005 and 1.0, respectively. 
We use ReLU activation function in all layers 
and set our batch size to 256. 
We train our model until convergence or up to 500 epochs 
with the Adam optimizer~\cite{kingma2014adam}.


\subsection{Overall Performance }
In TABLE \ref{tab:overall_acc}, we evaluate 
$\Design$ in low-resource scenarios using only $20\%$ of the training data. 
We regard each group of subjects, denoted as $\{T_0, T_1, \ldots\}$, as a target domain and the remaining groups as source domains. 
We repeat each experiment 3 times with 3 different random seeds and report the mean and standard deviation of LODO classification accuracy.   
$\Design$ exhibits notably higher accuracy and 
smaller standard deviations compared to SOTA approaches in low resource scenarios, indicating better performance and consistency.  

\subsection{Robustness in Low-Resource Scenarios}
We evaluate the robustness of $\Design$ by 
training our model with varying proportions of training data, i.e., $\{ 100\%, 80\%, 60\%,40\%, 20\% \}$ of data samples from source domains, and evaluating the trained model with the target domain. 
For each dataset, we report the average performance of our model on different target domains for each training data proportion in TABLE \ref{tab:partial_acc}.
With the decreasing amount of training data, while all models show performance degradation, $\Design$ consistently exhibits notably better performance compared to existing works. 
This indicates $\Design$ is considerably more robust to tackle DS across low-resource scenarios.

\section{Conclusion}
We propose $\Design$, a novel transformer-based contrastive meta-learning approach to achieve robust and generalizable  HAR  in low-resource distribution shift scenarios, outperforming SOTA algorithms
across various low-resource scenarios. 

\clearpage
\bibliographystyle{unsrt}
\bibliography{iccasp25}
\end{document}